\documentclass[11pt,a4paper]{article}
\usepackage[hyperref]{acl2021}
\usepackage{times}
\usepackage{latexsym}

\usepackage{microtype}

\usepackage{graphicx}
\usepackage{booktabs}
\usepackage{subfigure}
\usepackage{amsmath}
\usepackage{float}

\aclfinalcopy % Uncomment this line for the final submission
\def\aclpaperid{1643} %  Enter the acl Paper ID here

\title{Importance-based Neuron Allocation \\ 
for Multilingual Neural Machine Translation}

\author{Wanying Xie\textsuperscript{\rm 1,2,3}~~~Yang Feng\textsuperscript{\rm 1,2}\thanks{~ Corresponding author: Yang Feng. \newline \indent ~~ Our code can be got at https://github.com/ictnlp/NA-MNMT}~~~Shuhao Gu\textsuperscript{\rm 1,2}~~~Dong Yu\textsuperscript{\rm 3} \\ 
{\textsuperscript{\rm 1} {Key Laboratory of Intelligent Information Processing}} \\ Institute of Computing Technology, Chinese Academy of Sciences (ICT/CAS) \\
{ \textsuperscript{\rm 2} {University of Chinese Academy of Sciences, Beijing, China}} \\
{ \textsuperscript{\rm 3} Beijing Language and Culture University, China} \\
xiewanying07@gmail.com,~~yudong@blcu.edu.cn\\
\{{fengyang, gushuhao19b}\}@ ict.ac.cn
}

\date{}

\begin{document}
\maketitle
\begin{abstract}
Multilingual neural machine translation with a single model has drawn much attention due to its capability to deal with multiple languages. However, the current multilingual translation paradigm often makes the model tend to preserve the general knowledge, but ignore the language-specific knowledge.  Some previous works try to solve this problem by adding various kinds of language-specific modules to the model, but they suffer from the parameter explosion problem and require specialized manual design. To solve these problems, we propose to divide the model neurons into general and language-specific parts based on their importance across languages. The general part is responsible for preserving the general knowledge and participating in the translation of all the languages, while the language-specific part is responsible for preserving the language-specific knowledge and participating in the translation of some specific languages. Experimental results on several language pairs, covering IWSLT and Europarl corpus datasets, demonstrate the effectiveness and universality of the proposed method.
%Multilingual neural machine translation with a single model has drawn much attention due to its capability to deal with multiple languages. 
%However, this current multilingual translation paradigm often results in a disadvantage in language-specific knowledge due to a single translation model not being able to accommodate different languages in its limited parameter space. %说忽视language-specific是对的，但是最后说是因为limited parameter space不对，因为你的方法也是limited parameter space。
%Previous work to gain language-specific knowledge often adds special modules, which often increases the amount of model parameters and neurons. %再加上一个缺点，需要细致的手工调整。
%In response to these problems, we proposed an importance-based neuron allocation method to improve the overall performance without increasing model capacity. 
%Our approach seeks to retain as much of the neurons sharing generalization of translation models as possible, while still allowing for language-specific specialization of the other neurons to a particular language pair. %我们提出基于重要性将神经元分为两部分，shared神经元用于保留language-invariant知识，speicific的神经元用于保留speific的知识。
%Experimental results with Transformer on several language pairs, covering IWSLT and Europarl corpus datasets, demonstrate the effectiveness and universality of the proposed approach.
\end{abstract}

\section{Introduction}
%M-NMT因为可以用一个模型解码很多语言所以重要。
Neural machine translation(NMT)~\cite{DBLP:conf/emnlp/KalchbrennerB13,DBLP:conf/nips/SutskeverVL14,DBLP:journals/corr/BahdanauCB14,DBLP:conf/icml/GehringAGYD17,DBLP:conf/nips/VaswaniSPUJGKP17} has shown its superiority and drawn much attention in recent years. 
%There are thousands of languages in the world, and in practical application, it often faces the scenario of multiple languages translating with each other~\cite{DBLP:conf/emnlp/TanCHXQL19,DBLP:conf/naacl/AharoniJF19,DBLP:journals/corr/abs-1907-05019}. 
Although the NMT model can achieve promising results for high-resource language pairs, it is unaffordable to train separate models for all the language pairs since there are thousands of languages in the world~\citep{DBLP:conf/emnlp/TanCHXQL19,DBLP:conf/naacl/AharoniJF19,DBLP:journals/corr/abs-1907-05019}. 
%for a high-resource language pair can achieve promising results, it is unaffordable to train separate models for all the language pairs.
A typical solution to reduce the model size and the training cost is to handle multiple languages in a single multilingual neural machine translation (MNMT) model~\cite{DBLP:journals/corr/HaNW16,DBLP:conf/naacl/FiratCB16,DBLP:journals/tacl/JohnsonSLKWCTVW17,DBLP:conf/naacl/GuHDL18}. %这段可以简化，直接提出M-NMT的概念。%简化后
The standard paradigm of MNMT proposed by ~\newcite{DBLP:journals/tacl/JohnsonSLKWCTVW17} contains a language-shared encoder and decoder with a special language indicator in the input sentence to determine the target language. 
%The most popular paradigm of MNMT is proposed by ~\newcite{DBLP:journals/tacl/JohnsonSLKWCTVW17}, which contains one encoder and one decoder, where the encoder presents multiple source languages input and the decoder generates output for different target languages. During the training process, the training data from different languages are just combined together and the model is just trained with the mixed data. Besides, a special language indicator is added to the input sentence to determine the target language. 

%Unfortunately, it has been empirically shown that language signals obtained from language indicators alone are insufficient~\citep{DBLP:journals/corr/abs-1907-05019}.
Because different languages share all of the model parameters in the standard MNMT model, the model tends to converge to a region where there are low errors for all the languages. Therefore, the MNMT model trained on the combined data generally captures the general knowledge, but ignores the language-specific knowledge, rendering itself sub-optimal for the translation of a specific language ~\citep{DBLP:conf/wmt/SachanN18,DBLP:conf/coling/BlackwoodBW18,DBLP:conf/aaai/WangWSLT20}.
%Unfortunately, the standard MNMT model trained on the combined data generally tends to capture the language-invariant knowledge, but to ignore the language-specific knowledge because all of the parameters are shared across languages. rendering them sub-optimal for multilingual translation~\citep{DBLP:conf/wmt/SachanN18,DBLP:conf/coling/BlackwoodBW18}. It has been empirically shown that language signals obtained from language indicators alone are insufficient~\citep{DBLP:journals/corr/abs-1907-05019}.
%which necessitates an architecture dedicated to language-specific modeling~\citep{DBLP:conf/coling/BlackwoodBW18,DBLP:conf/wmt/SachanN18,zhang-etal-2020-improving,zhang2021share}.
%Towards learning a unified MNMT model, several researchers turn to augment the NMT model to learn language-specific knowledge. For example, ~\citet{DBLP:conf/coling/BlackwoodBW18} proposed a language-specific attention approach to translation model. ~\citet{vazquez-etal-2019-multilingual} and ~\citet{DBLP:journals/corr/abs-2004-06575} decoupled multilingual encoders and/or decoders to express more information. ~\citet{bapna-firat-2019-non} inserted lightweight adapters for each language pairs in encoder and decoder to increase its flexibility. However, while preserving the general knowledge, all the approaches require extra parameters and neurons to embed the particular knowledge at the same time. %介绍别人的工作可以更精简一些，多分析一些这些工作的缺点。1、参数爆炸 2、需要精细的手工调整
%上段修改后
To retain the language-specific knowledge, some researches turn to augment the NMT model with language-specific modules, e.g., the language-specific attention module~\citep{DBLP:conf/coling/BlackwoodBW18}, decoupled multilingual encoders and/or decoders~\citep{vazquez-etal-2019-multilingual,DBLP:journals/corr/abs-2004-06575} and the lightweight language adapters~\citep{DBLP:conf/emnlp/BapnaF19}. 
However, these methods suffer from the parameter increment problem, because the number of parameters increases linearly with the number of languages. Besides, the structure, size, and location of the module have a large influence on the final performance, which requires specialized manual design. As a result, these problems often prevent the application of these methods in some scenarios.
%However, while preserving the general knowledge, all the approaches require extra parameters and neurons to embed the particular knowledge at the same time. Besides, It requires fine manual adjustments to determine which layer adds which module, or to determine whether to specialize the attention module or an entire layer of encoder/decoder for better results.

%%基于以上，我们计划提出一个模型，既能够保留两种特征，也能够保证模型大小不变，同时免于繁琐的手工设计。To achieve this， 我们提出基于重要性将神经元分为两部分，shared神经元用于保留language-invariant知识，speicific的神经元用于保留speific的知识。
%tiao zheng hou
Based on the above, we aim to propose a method that can retain the general and language-specific knowledge, and keep a stable model size as the number of language-pair increases without introducing any specialized module. To achieve this, we propose to divide the model neurons into two parts based on their importance: the general neurons which are used to retain the general knowledge of all the languages, and the language-specific neurons which are used to retain the language-specific knowledge.
%In this paper we adopt an importance-based neuron allocation method without parameter increment to obtain both general representations and specific representations simultaneously. ~\citet{DBLP:conf/iclr/BauBSDDG19} and ~\citet{voita-etal-2019-analyzing} show that some important neurons in a well-trained NMT model are responsible for generating the translation and unimportant neurons can be erased without affecting the translation quality too much. According to these findings, we can preserve important neurons for all languages, while specializing other neurons for specific language pair.
%我们首先pretrain模型，然后我们使用重要性来当作分配神经元的标准，inspired by the work of ~。最后我们再fine-tune模型。
%gai
Specifically, we first pre-train a standard MNMT model on all language data and then evaluate the importance of each neuron in each language pair. According to their importance, we divide the neurons into the general neurons and the language-specific neurons. After that, we fine-tune the translation model on all language pairs. In this process, only the general neurons and the corresponding language-specific neurons for the current language pair participate in training. 
%More specifically, we start with a standard pre-trained MNMT model on all language data and then evaluate the importance of each neuron in each language pair. According to importance value, we allocate neurons into general neurons and language-specific neurons. After that, we fine-tune the translation model on all language pairs. In this process, only general neurons and specific neurons for the current language pair participate in training. 
%More specifically, we start with a standard pre-trained MNMT model on all language data and then evaluate the importance of each neuron in each language pair. According to the importance value, we preserve neurons that are important in all language pairs, which are thought to be responsible for general knowledge across languages(call them general neurons). For other neurons that are only important for certain language pairs, we allocate this neuron to its relevant language pairs(call them language-specific neurons). After this, we fine-tune the translation model on all language pairs, and only general neurons and specific neurons for this language pair on current language pair data participate in training. 
Experimental results on different languages show that the proposed method outperforms several strong baselines.

%Our main findings are summarized below:
Our contributions can be summarized as follows:
\begin{itemize}
  %\item We propose a method that can improve the translation performance of the MNMT model without increment of model capacity and fine manual design.
  \item We propose a method that can improve the translation performance of the MNMT model without introducing any specialized modules or adding new parameters.
  \item We show that the similar languages share some common features that can be captured by some specific neurons of the MNMT model.
  %\item We prove that other neurons except importance to all languages in the MNMT model can be reutilized to improve the specific language translation quality.
  %\item Our model can keep superior performance over baselines on basis of no increment of model capacity.
  \item We show that some modules tend to capture the general knowledge while some modules are more essential for capturing the language-specific knowledge.
  %\item The importance-based neuron allocation method achieves a hierarchical neuron sharing mode so that each neuron is only responsible for the language pairs associated with it.
\end{itemize}

\section{Background}

In this section, we will give a brief introduction to the Transformer model~\citep{DBLP:conf/nips/VaswaniSPUJGKP17} and the Multilingual translation.

\subsection{The Transformer}

%We apply our method in the framework of \textsc{Transformer}~\cite{DBLP:conf/nips/VaswaniSPUJGKP17} which will be briefly introduced here. 
We denote the input sequence of symbols as $\mathbf{x'}=(x_1,\ldots,x_J)$, the ground-truth sequence as $\mathbf{y}^{*}=(y_1^{*},\ldots,y_{K*}^{*})$ and the translation as $\mathbf{y}=(y_1,\ldots,y_K)$.%

%\paragraph{The Encoder \& Decoder:}
Transformer is a stacked network with $\mathnormal{N}$ identical layers containing two or three basic blocks in each layer. For a single layer in the encoder, it consists of a multi-head self-attention and a position-wise feed-forward network. For a single decoder layer, besides the above two basic blocks, a multi-head cross-attention follows multi-head self-attention. The input sequence $\mathbf{x}$ will be first converted to a sequence of vectors and fed into the encoder. Then the output of the $\mathnormal{N}$-th encoder layer will be taken as source hidden states and fed into decoder. The final output of the $\mathnormal{N}$-th decoder layer gives the target hidden states and translate the target sentences.

%\paragraph{The Objective} 
%The model is optimized by minimizing a cross-entropy loss of the ground-truth sequence with teacher forcing training:
%\begin{equation}\label{eq::loss}
%    \mathcal{L}(\theta) = -\frac{1}{K} \sum_{k=1}^{K} \log p(y_k^{*} | \mathbf{y}_{<k}, \mathbf{x}; \theta),
%\end{equation}
%where $K$ is the length of the target sentence and $\theta$ denotes the model parameters.

\subsection{Multilingual Translation}

%Multilingual models perform the multi-task paradigm with some degree of parameter sharing,
In the standard paradigm of MNMT, all parameters are shared across languages and the model is jointly trained on multiple language pairs. We follow ~\citet{DBLP:journals/tacl/JohnsonSLKWCTVW17} to reuse standard bilingual NMT models for multilingual translation by altering the source input with a language token \emph{lang}, i.e. changing $\mathbf{x'}$ to $\mathbf{x} = ({\emph{lang}, x_1,\ldots,x_J})$.

\begin{figure*}[t!]
    \centering
    \includegraphics[width=2.0\columnwidth]{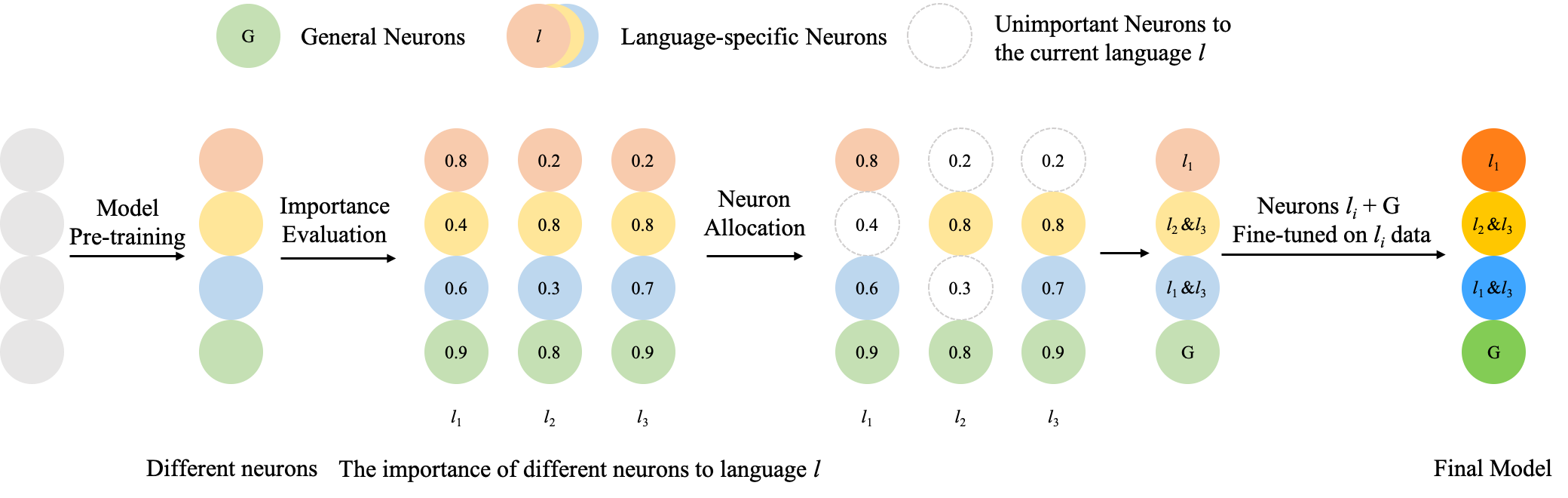}
    \caption{The whole training process of the proposed method. The red, yellow and blue circles represent language-specific neurons that are important for $l_1$, $l_2$\&$l_3$ and $l_1$\&$l_3$, respectively.}
    %\caption{The whole training process of the proposed method. The green circles represent the general neurons which are important for all languages. The red, yellow and blue circles represent language-specific neurons that are important for $l_1$, $l_2$\&$l_3$ and $l_1$\&$l_3$, respectively. The grey circles represent unimportant neurons to the current language $l$.}
    \label{fig:approach}
\end{figure*}

\section{Approach}

Our goal is to build a unified model, which can achieve good performance on all language pairs. The main idea of our method is that different neurons have different importance to the translation of different languages. Based on this, we divide them into general and language-specific ones and make general neurons participate in the translation of all the languages while language-specific neurons focus on some specific languages. Specifically, the proposed approach involves the following steps shown in Figure~\ref{fig:approach}. First, we pretrain the model on the combined data of all the language pairs following the normal paradigm in~\citet{DBLP:journals/tacl/JohnsonSLKWCTVW17}. Second, we evaluate the importance of different neurons on these language pairs and allocate them into general neurons and language-specific neurons. Last, we fine-tune the translation model on the combined data again. It should be noted that for a specific language pair only the general neurons and the language-specific neurons for this language pair will participate in the forward and backward computation when the model is trained on this language pair. Other neurons will be zeroed out during both training and inference.
%When trained on data of language pair $lang$, only general neurons and language-specific neurons for $lang$ are participate forward and backward computation.

\subsection{Importance Evaluation}

The basic idea of importance evaluation is to determine which neurons are essential to all languages while which neurons are responsible for some specific languages. 
For a neuron $i$, its average importance $\mathcal{I}$ across language pairs is defined as follow:
%For the set of neurons ${I}=(i_1,\ldots,i_K)$, the neurons' overall importance for all languages $\mathcal{I}$ is formulated below:
\begin{equation}\label{eq::ivb}
    \mathcal{I}(i) = \frac{1}{M} \sum_{m=1}^{M} \Theta^m(i),
\end{equation}
where the $\Theta(\cdot)$ denotes the importance evaluation function and $M$ denotes the number of language pairs. This value correlates positively with how important the neuron is to all languages.
%The larger the value is, the more important the neuron is for all languages. 
For the importance evaluation function $\Theta(\cdot)$, we adopt two schemes: one is based on the Taylor Expansion and the other is based on the Absolute Value.

\paragraph{Taylor Expansion} 
We adopt a criterion based on the Taylor Expansion~\citep{DBLP:conf/iclr/MolchanovTKAK17}, where we directly approximate the change in loss when removing a particular neuron. Let $h_i$ be the output produced from neuron $i$ and $H$ represents the set of other neurons. 
 %Because we have already assumed the independence of the neurons in the back propagation, 
Assuming the independence of each neuron in the model, the change of loss when removing a certain neuron can be represented as:
\begin{equation}
    |\Delta\mathcal{L}(h_i)| = |\mathcal{L}(H, h_i=0) - \mathcal{L}(H, h_i)|,
\end{equation} 
where $\mathcal{L}(H, h_i=0)$ is the loss value if the neuron $i$ is pruned and $\mathcal{L}(H, h_i)$ is the loss if it is not pruned. For the function $\mathcal{L}(H, h_i)$, its Taylor Expansion at point $h_i = a$ is:
\begin{equation}
  \mathcal{L}(H, h_i) = \sum_{n=0}^{N}\frac{\mathcal{L}^{n}(H, a)}{n!}(h_i - a)^n + R_N(h_i),
\end{equation}
where $\mathcal{L}^{n}(H, a)$ is the $n$-th derivative of $\mathcal{L}(H, h_i)$ evaluated at point $a$ and $R_N(h_i)$ is $N$-th remainder. 
Then, approximating $\mathcal{L}(H, h_i=0)$ with a first-order Taylor polynomial where $h_i$ equals zero:
\begin{equation}
    \mathcal{L}(H, h_i=0) = \mathcal{L}(H, h_i)-\frac{\partial \mathcal{L}(H, h_i)}{\partial h_i}h_i-R_1(h_i).
\end{equation}
The remainder $R_1$ can be represented in the form of Lagrange:
\begin{equation}
    R_1(h_i) = \frac{\partial^2\mathcal{L}(H, h_i)}{\partial^2\delta h_i}h_i^2,
\end{equation}
where $\delta \in (0,1)$. Considering the use of ReLU activation function~\citep{DBLP:journals/jmlr/GlorotBB11} in the model, the first derivative of loss function tends to be constant, so the second order term tends to be zero in the end of training. Thus, we can ignore the remainder and get the importance evaluation function as follows:
\begin{equation}
    \Theta_\mathrm{TE}(i) = \left|\Delta\mathcal{L}(h_i)\right|  =  \left|\frac{\partial \mathcal{L}(H, h_i)}{\partial h_i}h_i\right|.
\end{equation} 
%Intuitively, this criterion prunes neurons that have an almost flat gradient of the objective function. 
In practice, we need to accumulate the product of the activation and the gradient of the objective function w.r.t to the activation, which is easily computed during back-propagation. Finally, the evaluation function is shown as:
\begin{equation}\label{eq:te}
    \Theta^{m}_\mathrm{TE}(i^l)=\frac{1}{T_m}\sum_{t}\left|\frac{\delta\mathcal{L}(H, h_i^l)}{\delta h_i^l}h_i^l\right|,
\end{equation}
where $h_i^l$ is the activation value of the $i$-th neuron of $l$-th layer and $T_m$ is the number of the training examples of language pair $m$. The criterion is computed on the data of language pair $m$ and averaged over $T_m$. %Finally, neurons are distinguished according to their importance computed by $\Theta_\mathrm{TE}$.
 
\paragraph{Absolute Value} 
We adopt the magnitude-based neuron importance evaluation scheme~\citep{DBLP:conf/conll/SeeLM16}, where the absolute value of each neuron's activation value is treated as the importance:
\begin{equation}
\label{eq:av}
    \Theta^m_\mathrm{AV}(i^l)=\frac{1}{T_m}\sum_{t}|h_i^l|.
\end{equation}
The notations in the above equation are the same as those in the Equation~\ref{eq:te}. After the importance of each neuron is evaluated on the combined data, we need to determine the role of each neuron in the fine-tuning step following the method in the next section.
%where $h_i^l$ is the activation value of the $i$-th neuron and $T_m$ is the amount of the training examples of language pair $lang$. 
%Finally, a certain percentage $\rho$ of neurons in each module are regard as general neurons.

\subsection{Neuron Allocation}

%The main idea of allocation neurons is distinguishing different neurons into general neurons and language-specific neurons, and make general neurons compromise between language pairs while language-specific neurons focus on their relevant language pairs. 

In this step, we should determine which neurons are shared across all the language pairs and which neurons are shared only for some specific language pairs. 
%Other variants are based on source language, target language, and so on. We demonstrate in Appendix~\ref{variet} that our approach is superior to the other variants.

%Appendix~\ref{variet}, we verify our method better than different varieties

\paragraph{General Neurons}
According to the overall importance $\mathcal{I}(i)$ in Equation~\ref{eq::ivb}, the value correlates positively with how important the neuron is to all languages. Therefore, we rank the neurons in each layer based on the importance and make the top $\rho$ percentage as general neurons that are responsible for capturing general knowledge.

\paragraph{Language-specific Neurons} Next, we regard other neurons except for the general neurons as the language-specific neurons and determine which language pair to assign them to. To achieve this, we compute an importance threshold for each neuron:
\begin{equation}
\label{eq:kmax}
\begin{split}
     \lambda(i) &= k \times \max(\Theta^m(i)), \\
    & m \in \{1,\ldots, M\}, k \in [0, 1]
\end{split}
\end{equation}
, where $\max(\Theta^m(i))$ denotes the maximum importance of this neuron in all language pairs and $k$ is a hyper-parameter. The neuron will be assigned to the language-pairs whose importance is larger than the threshold. When the importance of neurons is determined, the number of language pairs associated with each neuron can be adjusted according to $k$. The smaller the $k$, the more language-pairs will be associated with the specific neurons.
%The language pair whose importance is larger than the threshold can take possession of this neuron. 
%For example, see yellow circle in Figure~\ref{fig:approach}. If set k as 0.7 then $\lambda = k \times max = 0.7 * 0.8 = 0.56$. The importance values of language pair $l_2$ and $l_3$ are greater than $\lambda = 0.56$, so this neuron is language-specific neurons to the two language pairs.
%The threshold $\lambda$ is controlled by adjusting the $k$ value. 
%If $k$ is larger, then a neuron will be responsible for fewer language pairs, and therefore the neuron is more specific. Conversely, if $k$ is smaller, then a neuron is responsible for more language pairs, and therefore the neuron is more sharing.
In this way, we flexibly determine the language pairs assigned to each neuron according to its importance in different languages. Note that the neuron allocation is based on the importance of \emph{language pair}. 
We have also tried other allocation variants, e.g., based on the source language, target language, and find that the language pair-based method is the best among of these methods. The detailed results are listed in Appendix~\ref{variet}.

After this step, the model is continually fine-tuned on the combined multilingual data. If the training data is from a specific language pair, only the general neurons and the language-specific neurons for this language pair will participate in the forward computation and the parameters associated with them will be updated during the backward propagation. 

%After this step, we dynamically adjust the language pairs assigned to each neuron according to its importance in different languages.

%At last, the general neurons are shared across all language pairs, while the language-specific neurons are shared to part of language pairs or a particular language pair. Through this neuron allocation method, we make each neuron flexibly determine the relevant language pair according to its own importance situation, and realize the \emph{hierarchical} sharing mode.

\section{Experiments}

\subsection{Data Preparation}

In this section, we describe the datasets using in our experiments on many-to-many and one-to-many multilingual translation scenarios.

\paragraph{Many-to-Many} For this translation scenario, we test our approach on IWSLT-17\footnote{https://sites.google.com/site/iwsltevaluation2017} translation datasets, including English, Italian, Romanian, Dutch (briefly, En, It, Ro, Nl). We experimented in eight directions, including It$\leftrightarrow$En, Ro$\leftrightarrow$En, Nl$\leftrightarrow$En, and It$\leftrightarrow$Ro, with 231.6k, 220.5k, 237.2k, and 217.5k data for each language pair. We choose test2016 and test2017 as our development and test set, respectively. Sentences of all languages were tokenized by the Moses scripts\footnote{http://www.statmt.org/moses/} and further segmented into subword symbols using Byte-Pair Encoding (BPE) rules~\citep{DBLP:conf/acl/SennrichHB16a} with 40K merge operations for all languages jointly.

\paragraph{One-to-Many} We evaluate the quality of our multilingual translation models using training data from the Europarl Corpus\footnote{http://www.statmt.org/europarl/}, Release V7. Our experiments focus on English to twelve primary languages: Czech, Finnish, Greek, Hungarian, Lithuanian, Latvian, Polish, Portuguese, Slovak, Slovene, Swedish, Spanish (briefly, Cs, Fi, El, Hu, Lt, Lv, Pl, Pt, Sk, Sl, Sv, Es). For each language pair, we randomly sampled 0.6M parallel sentences as training corpus (7.2M in all). The Europarl evaluation data set dev2006 is used as our validation set, while devtest2006 is our test set. For language pairs without available development and test set, we randomly split 1K unseen sentence pairs from the corresponding training set as the development and test data respectively. We tokenize and truecase the sentences with Moses scripts and apply a jointly-learned set of 90k BPE obtained from the merged source and target sides of the training data for all twelve language pairs.

\begin{table*}
\centering
\resizebox{2.0\columnwidth}!{
\begin{tabular}{lllllllllll}
\bottomrule
 & It$\rightarrow$En & En$\rightarrow$It & Ro$\rightarrow$En & En$\rightarrow$Ro & Nl$\rightarrow$En & En$\rightarrow$Nl & It$\rightarrow$Ro & Ro$\rightarrow$It & AVE & Para \\ 
\hline
\hline
\textbf{Individual} & 34.99 & 31.22 & 28.58 & 23.19 & 30.21 & 27.69 & 19.52 & 20.95 & 27.04 & 466.4M \\ 
\hline
\textbf{Multilingual} & 37.55 & 32.62 & 31.58 & 24.64 & 31.13 & 28.86 & 20.82 & 23.79 & 28.87  & \textbf{64.69M}\\ 
\textbf{\quad +TS} & 38.11 & 33.46 & 31.82 & 24.96 & 32.04 & 30.06 & 21.43 & 23.59 & 29.43$^{+0.56}$  & 121.42M \\ 
\textbf{\quad +Adapter} & 38.25 & 34.16 & 32.07 & 25.08 & 32.56 & 29.66 & 21.18 & 24.26 & 29.65$^{+0.78}$  & 77.43M \\ 
\textbf{Our Method-AV} & 38.07 & 34.15 & 32.17 & 26.00 & 32.21 & 30.11 & 21.96 & 24.46 & 29.89$^{+1.02}$  & \textbf{64.69M} \\ 
\textbf{Our Method-TE} & \textbf{38.31} & \textbf{34.24} & \textbf{32.24} & \textbf{26.34} & \textbf{32.73} & \textbf{30.16} & \textbf{22.21} & \textbf{24.76} & \textbf{30.12}$^{+1.25}$  & \textbf{64.69M} \\ 
\bottomrule
\end{tabular}
}
\caption{BLEU scores on the many-to-many translation tasks. 'AVE' denotes the average BLEU of the eight test sets and 'Para' denotes the number of parameters of the whole model. 'Para' of the Individual system is the sum of the models for the eight language pairs with 58.3M parameters for each model.}% * and ** mean the improvements over the Multilingual method is statistically significant~\citep{DBLP:conf/acl/CollinsKK05} ($\rho < 0.05$ and $\rho < 0.01$, respectively).}
\label{tab:resultsmall}
\end{table*}

\begin{table*}
\centering
\resizebox{2.0\columnwidth}!{
\begin{tabular}{lllllllllllllll}
\bottomrule
 & Cs & El & Es & Fi & Hu & Lt & Lv & Pl & Pt & Sk & Sl & Sv & AVE & Para \\ 
\hline
\hline
\textbf{Individual} & 36.14 & 39.86 & 41.16 & 22.95 & 31.75 & 32.31 & 38.12 & 32.95 & 35.57 & 40.51 & 43.83 & 33.23 & 35.70  & 746.76M \\ 
\hline
\textbf{Multilingual} & 37.87 & 40.34 & 41.58 & 23.03 & 31.10 & 33.11 & 39.22 & 32.67 & 36.20 & 42.05 & 44.76 & 33.16 & 36.26  & \textbf{90.42M}\\ 
\textbf{\quad +TS} & 37.70 & 40.70 & 42.05 & 23.28 & 31.78 & 32.90 & 39.48 & 33.66 & 36.09 & 42.03 & 44.29 & 33.14 & 36.43$^{+0.17}$  & 273.77M \\ 
\textbf{\quad +Adapter} & 38.11 & 40.23 & 41.83 & 23.66 & 32.00 & 33.49 & 39.87 & 32.85 & 36.25 & 42.00 & 44.63 & 32.90 & 36.49$^{+0.23}$  & 109.54M \\ 
\textbf{Our Method-AV} & 37.84 & \textbf{40.75} & 42.16 & 23.71 & 31.40 & 33.56 & 39.95 & 33.23 & 36.56 & 42.09 & 45.27 & 33.38 & 36.66$^{+0.40}$  & \textbf{90.42M} \\ 
\textbf{Our Method-TE} & \textbf{38.21} & 40.70 & 42.22 & \textbf{23.74} & 31.32 & 33.55 & 39.78 & 32.94 & \textbf{36.58} & 41.91 & 44.94 & 33.07 & 36.58$^{+0.32}$  & \textbf{90.42M} \\ 
\textbf{\quad +Expansion} & 38.03 & 40.59 & \textbf{42.28} & 23.73 & \textbf{32.47} & \textbf{34.12} & \textbf{40.12} & \textbf{33.95} & 36.41 & \textbf{42.44} & \textbf{45.30} & \textbf{33.43} & \textbf{36.91}$^{+0.65}$ & 102.14M \\ 
\bottomrule
\end{tabular}
}
\caption{BLEU scores on one-to-many translation tasks. 'Para' of the Individual system is 62.23M for each language pair. The denotations represent the same meaning as in Table~\ref{tab:resultsmall}.}
\label{tab:resultbig}
\end{table*}

\subsection{Systems}

To make the evaluation convincing, we re-implement and compare our method with four baseline systems, which can be divided into two categories with respect to the number of models. The multiple-model approach requires maintaining a dedicated NMT model for each language:

\textbf{Individual} A NMT model is trained for each language pair. Therefore, there are $N$ different models for $N$ language pairs.

The unified model-based methods handle multiple languages within a single unified NMT model:

\textbf{Multilingual}~\citep{DBLP:journals/tacl/JohnsonSLKWCTVW17} Handling multiple languages in a single transformer model which contains one encoder and one decoder with a special language indicator \emph{lang} added to the input sentence.

\textbf{+TS}~\citep{DBLP:conf/coling/BlackwoodBW18} This method assigns language-specific attention modules to each language pair. We implement the target-specific attention mechanism because of its excellent performance in the original paper. 
%Task-specific attention models allows for language-specific specialization of the attention model to a particular language-pair or task~\citep{DBLP:conf/coling/BlackwoodBW18}. Our implementation was target-specific attention because of its excellent performance in the original paper.

\textbf{+Adapter}~\citep{DBLP:conf/emnlp/BapnaF19} This method injects tiny adapter layers for specific language pairs into the original MNMT model. We set the dimension of projection layer to 128 and train the model from scratch.

\textbf{Our Method-AV} Our model is trained just as the Approach section describes. In this system, we adopt the absolute value based method to evaluate the importance of neurons across languages. 
%On a well trained model we use absolute value for importance evaluation and apply the proposed neuron allocation approach.

\textbf{Our Method-TE} This system is implemented the same as the system \emph{Our Method-AV} except that we adopt the Taylor Expansion based evaluation method as shown in Equation~\ref{eq:te}.

\textbf{+Expansion} To make a fair comparison, we set the size of Feed Forward Network to 3000 to expand the model capacity up to the level of other baselines, and then apply our Taylor Expansion based method to this model.

%%%%%%%% importance distribution %%%%%%%%
\begin{figure}[htbp]
\centering
\subfigure[O2M-Enc-6-FFN]{
\label{importance1}
\includegraphics[width=0.8\columnwidth]{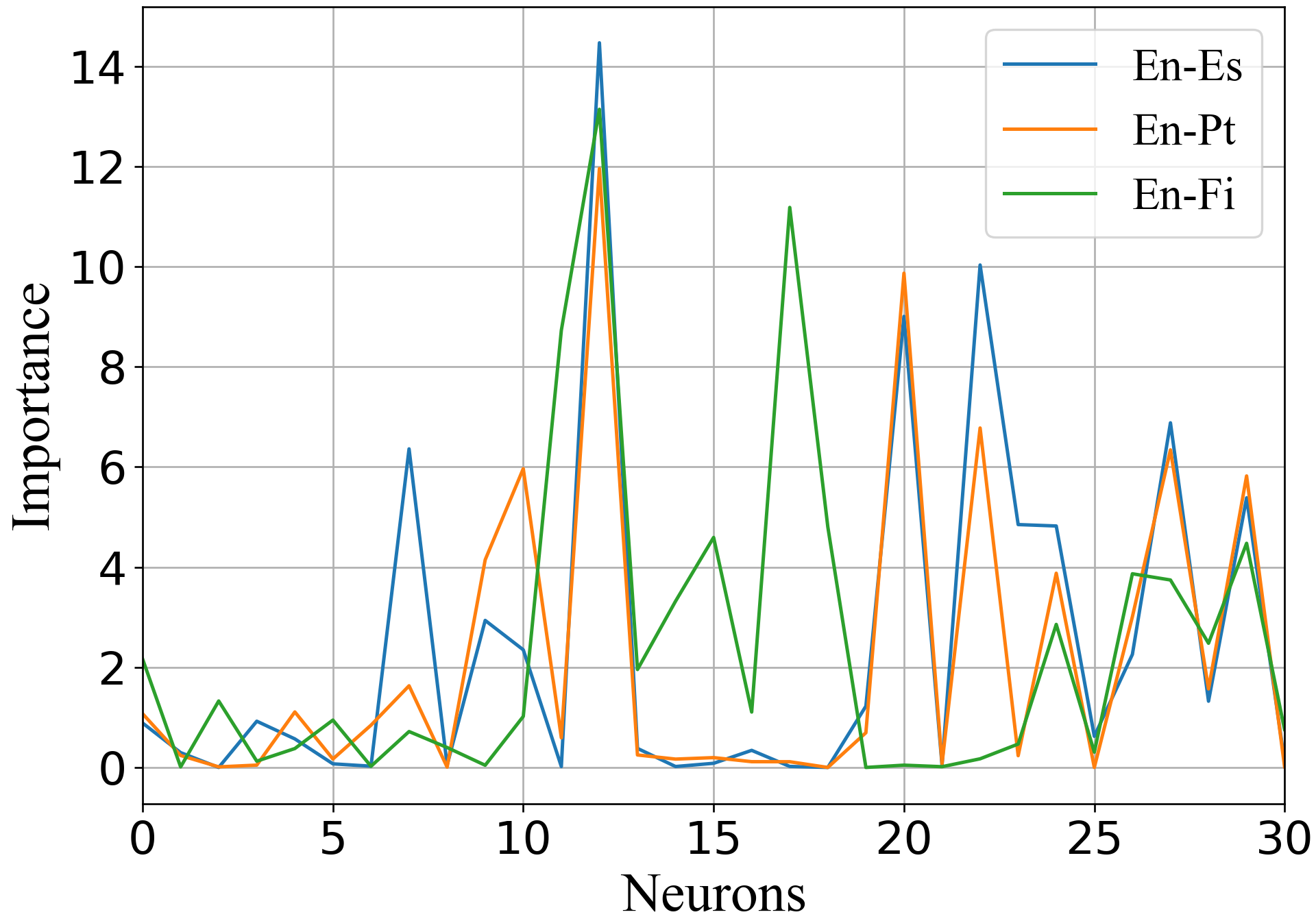}
}
\subfigure[O2M-Dec-6-FFN]{
\label{importance2}
\includegraphics[width=0.8\columnwidth]{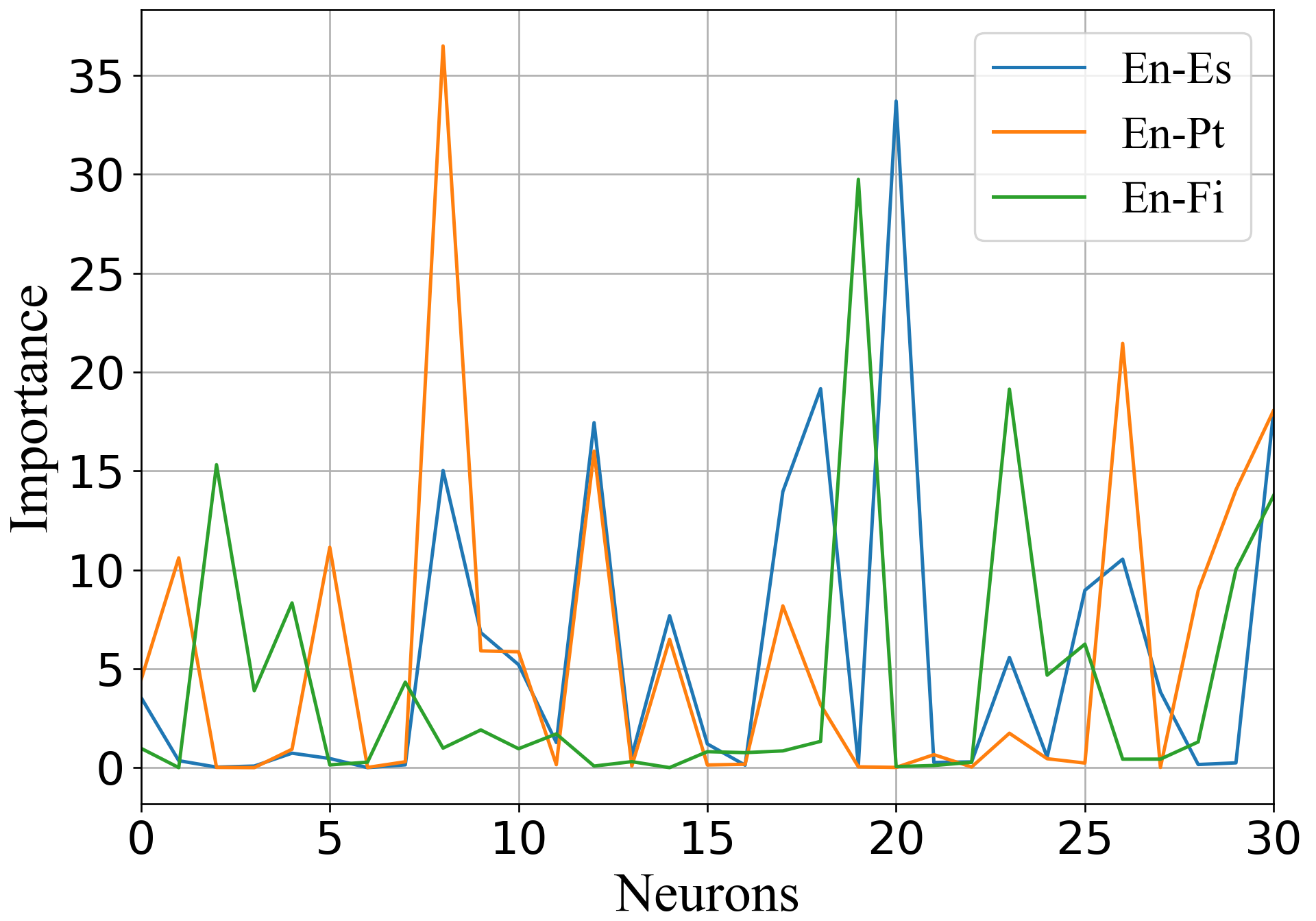}
}
\subfigure[M2M-Enc-6-FFN]{
\label{importance3}
\includegraphics[width=0.8\columnwidth]{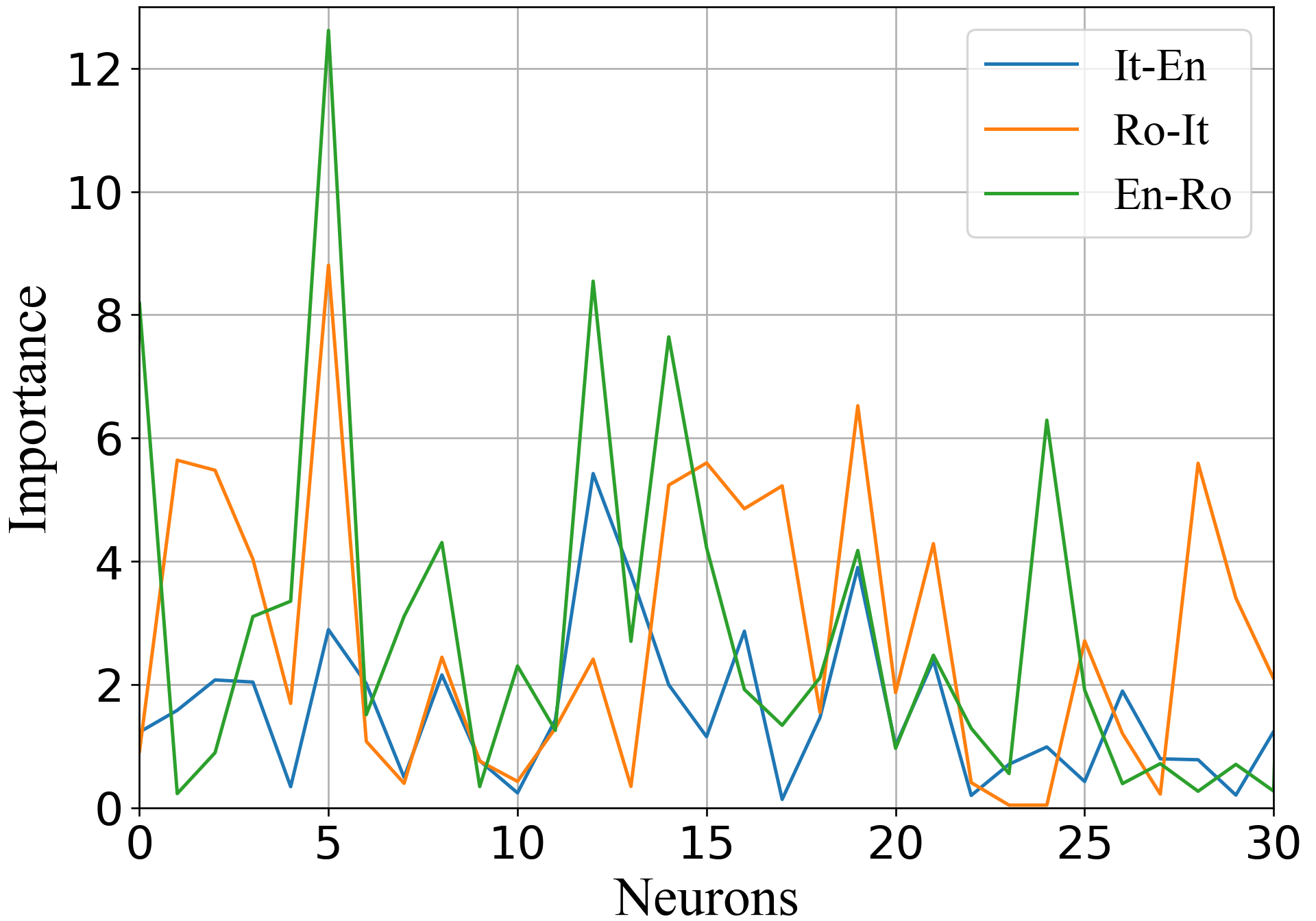}
}
\subfigure[M2M-Dec-6-FFN]{
\label{importance4}
\includegraphics[width=0.8\columnwidth]{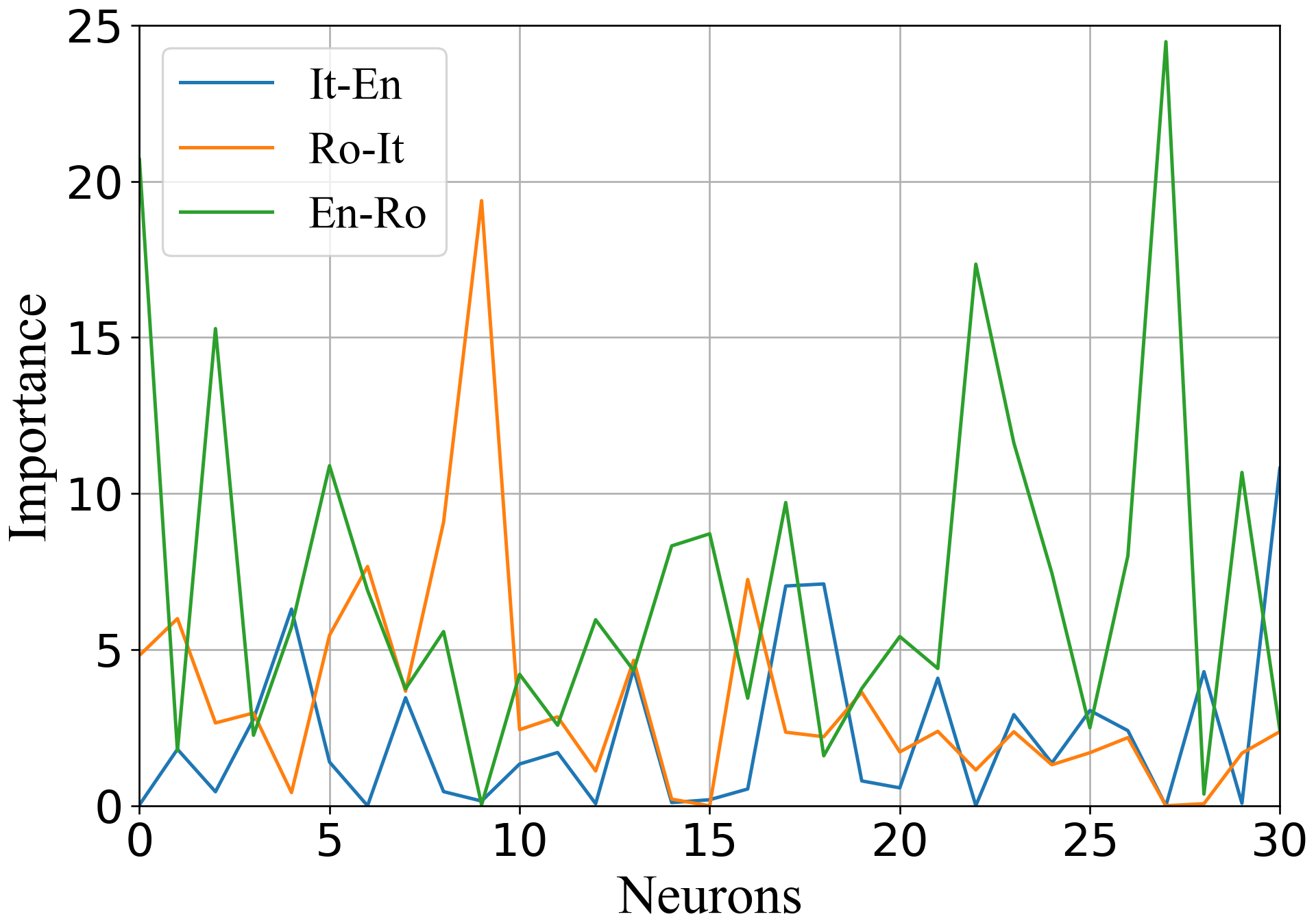}
}
\centering
\caption{Importance distribution of neurons computed by Taylor Expansion in each module. For example, 'O2M-Enc-6-FFN' represents the importance of the feed forward network in the 6-th encoder layer.}
\label{importance}
\end{figure}

\subsection{Details}
For fair comparisons, we implement the proposed method and other contrast methods on the advanced Transformer model using the open-source toolkit {\em Fairseq-py}~\citep{ott2019fairseq}. We follow ~\citet{DBLP:conf/nips/VaswaniSPUJGKP17} to set the configurations of the NMT model, which consists of 6 stacked encoder/decoder layers with the layer size being 512. All the models were trained on 4 NVIDIA 2080Ti GPUs where each was allocated with a batch size of 4,096 tokens for one-to-many scenario and 2,048 tokens for the many-to-many scenario. We train the baseline model using Adam optimizer~\citep{DBLP:journals/corr/KingmaB14} with $\beta_1=0.9$, $\beta_2=0.98$, and $\epsilon=10^{-9}$. The proposed models are further trained with corresponding parameters initialized by the pre-trained baseline model. 
We vary the hyper-parameter $\rho$ that controls the proportion of general neurons in each module from $80\%$ to $95\%$ and set it to $90\%$ in our main experiments according to the performance. The detailed results about this hyper-parameter are list in Appendix~\ref{app:rho}. We set the hyper-parameter $k$ to $0.7$ and do more analysis on it in Section~\ref{sec:kmax}.
%We set the hyper-parameters $\rho=90\%$ and $k=0.7$. Refer to Appendix~\ref{app:rho} and Section~\ref{sec:kmax} for the selection experiment of the hyper-parameters respectively. 
For evaluation, we use beam search with a beam size of 4 and length penalty $\alpha = 0.6$.

\subsection{Results}

The final translation is detokenized and then the quality is evaluated using the $4$-gram case-sensitive BLEU~\citep{DBLP:conf/acl/PapineniRWZ02} with the {\em SacreBLEU} tool~\citep{post-2018-call}.\footnote{BLEU+case.mixed+numrefs.1+smooth.exp+tok.13a +version.1.4.14} 
%We compare our method with the basic method which handles multiple languages in dedicated models and a uniform model. We also compare our method with methods using different approaches, such as ~\citet{DBLP:conf/coling/BlackwoodBW18} (named +TS) which allow proposed language-specific attention to a particular language-pair or task, and ~\citet{DBLP:conf/emnlp/BapnaF19} (named +Adapter) which inject tiny adapter layers for specific language pair. We also compared the two different importance evaluation methods previously mentioned TE and AV. 

\paragraph{Many-to-Many} The results are given in Table~\ref{tab:resultsmall}. We can see that the improvements brought by +TS and +Adapter methods are not large. For the +TS method, attention module may be not essential to capture language-specific knowledge, and thus it is difficult to converge to good optima. For the +Adapter method, adding an adapter module to the end of each layer may be not appropriate for some languages and hence has a loose capture to the specific features. In all language pairs, our method based on Taylor Expansion outperforms all the baselines in the datasets. Moreover, the parameters in our model are the same as the Multilingual system and less than other baselines.

\paragraph{One-to-Many} The results are given in Table~\ref{tab:resultbig}, our method exceeds the multilingual baseline in all language pairs and outperforms other baselines in most language pairs without capacity increment. When we expand the model capacity to the level of +Adapter, our approach can achieve better translation performance, which demonstrates the effectiveness of our method. Another finding is that the results of the individual baseline are worse than other baselines. The reason may be the training data is not big enough, individual baseline can not get a good enough optimization on 0.6M sentences, while the MNMT model can be well trained with a total of 7.2M data.
% total of 7.2M data can train a satisfactory MNMT model.
%Another finding is that all the methods cannot achieve big improvements on the one-to-many translation. The reason may be the training data is big enough. The multilingual baseline has already got a good enough optimization and it is difficult for further improvements.

%%%%%%%% neurons distribution %%%%%%%%
\begin{figure}[htbp!]
\centering
\subfigure[Encoder]{
\label{sumenc}
\includegraphics[width=0.8\columnwidth]{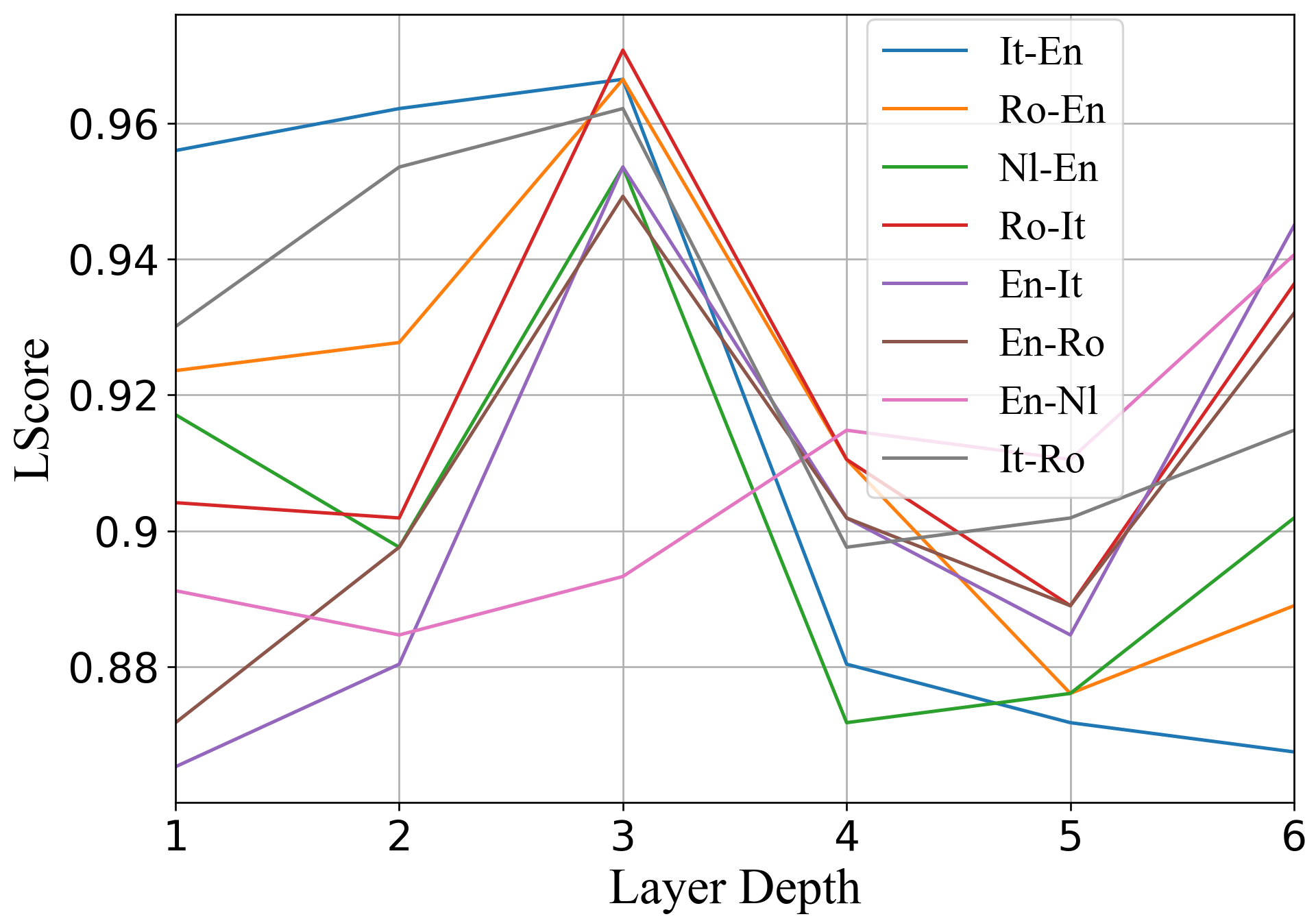}
}
\subfigure[Decoder]{
\label{sumdec}
\includegraphics[width=0.8\columnwidth]{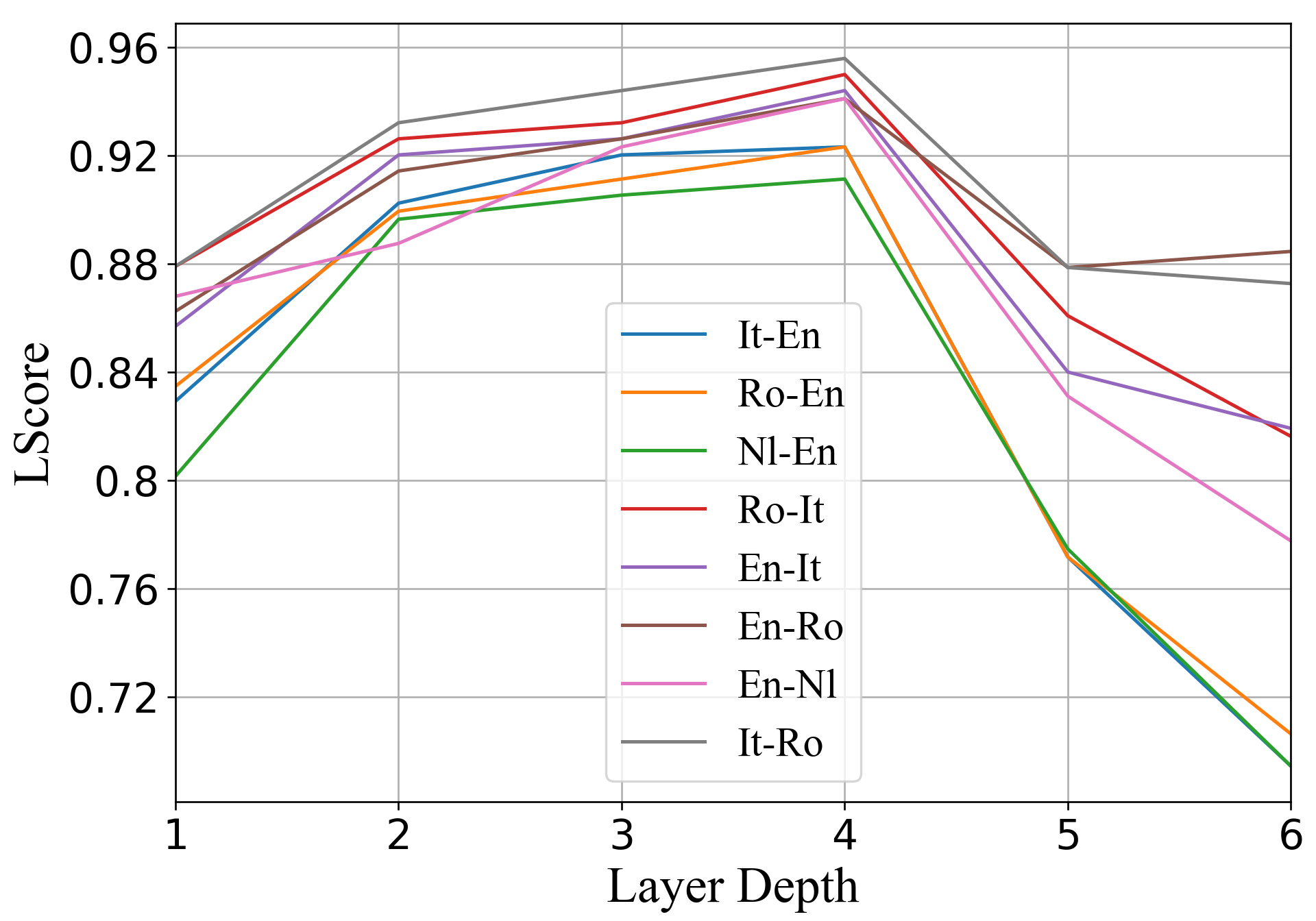}
}
\subfigure[Encoder]{
\label{2enc}
\includegraphics[width=0.8\columnwidth]{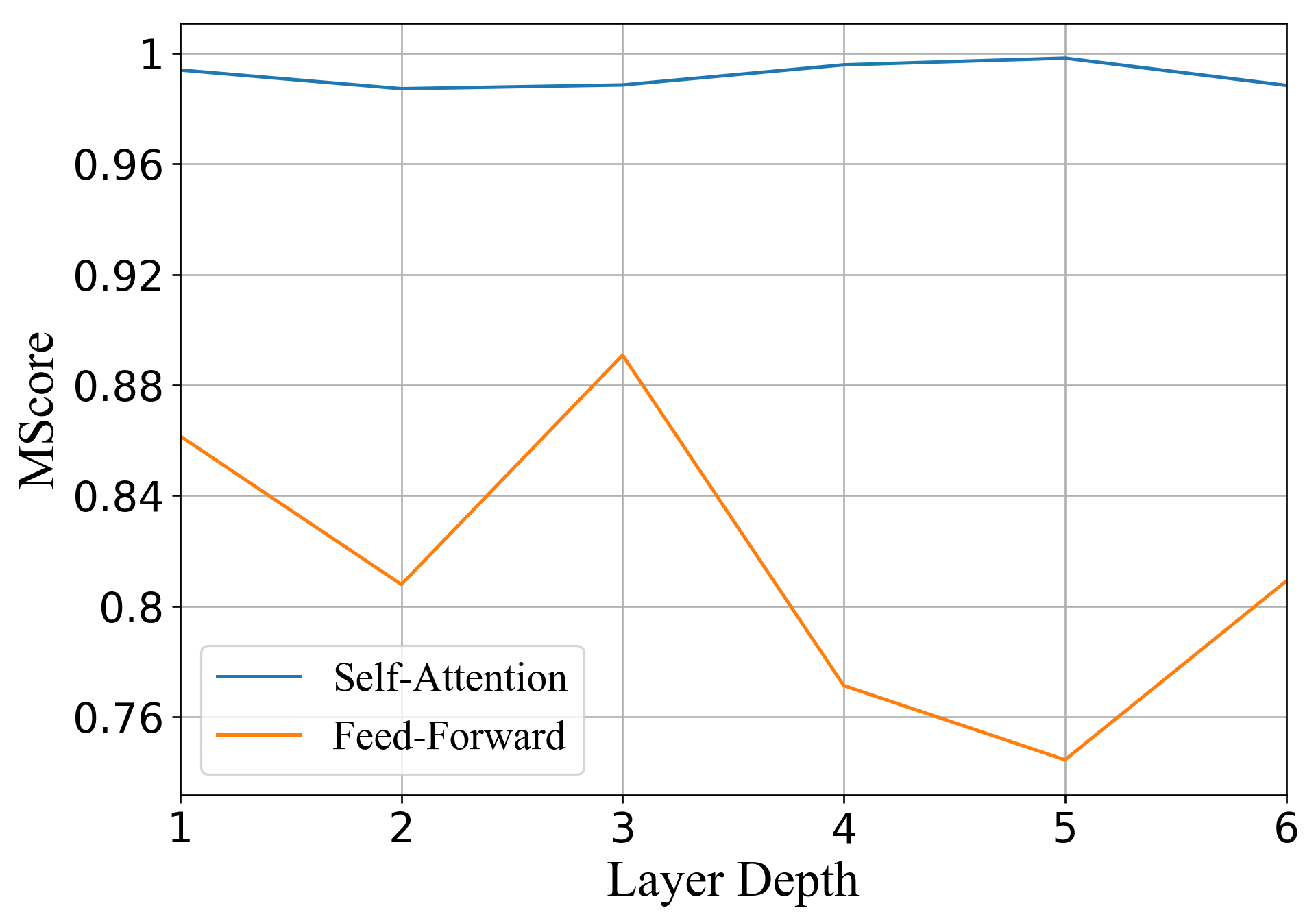}
}
\subfigure[Decoder]{
\label{3dec}
\includegraphics[width=0.8\columnwidth]{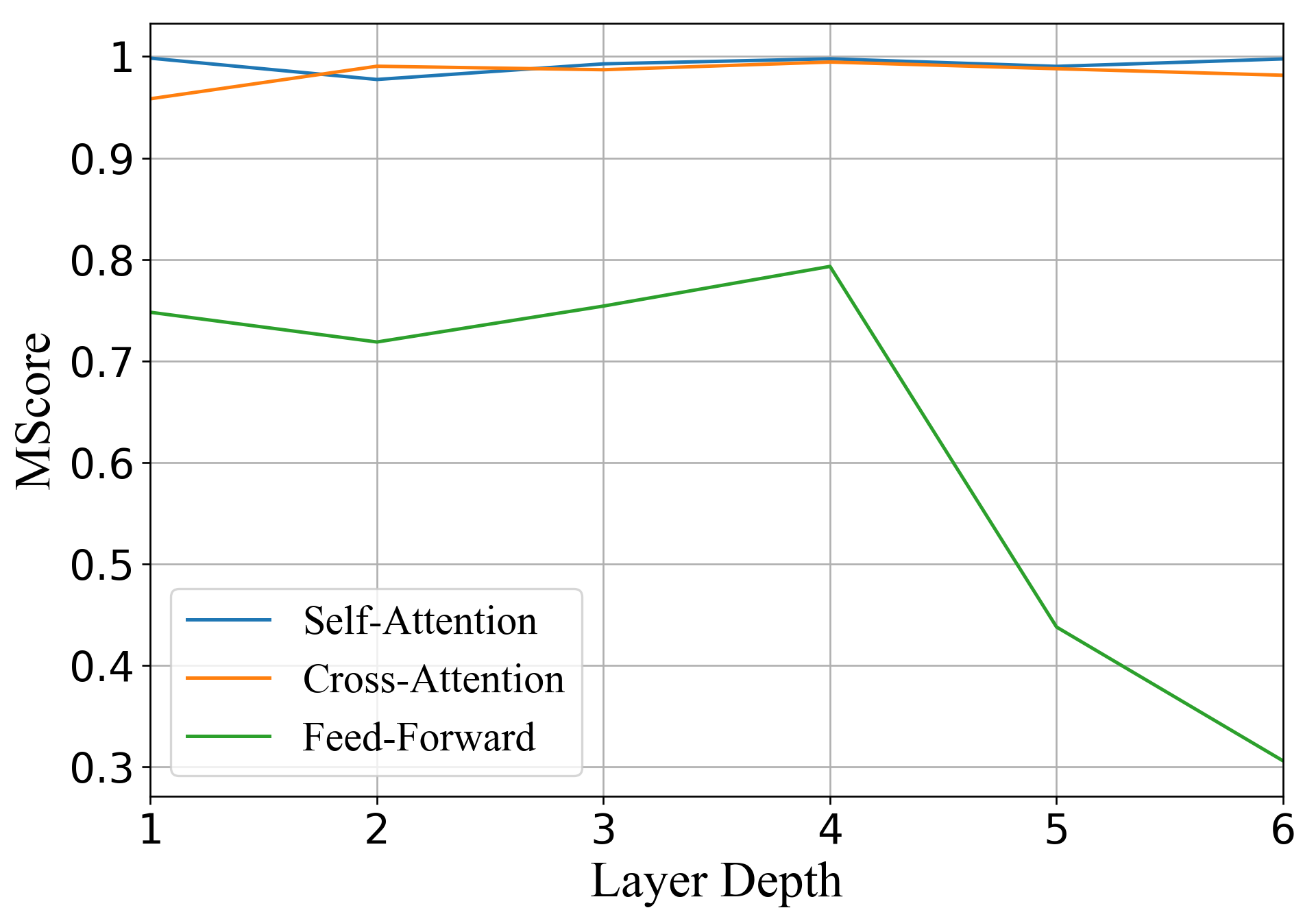}
}
\centering
\caption{The distribution of the language-specific neurons in the encoder and decoder. The importance of neurons is computed by Taylor Expansion. The first two sub-figures show the proportion of specific neurons for different language pairs, while the last two sub-figures show the proportion of specific neurons in different modules.}
\label{neurons}
\end{figure}

\section{Analysis}

%In the analysis section, we will first report the neuron importance for different languages, then analyze the distribution of different language pairs and modules in the model, then show the effects of hyper-parameter $k$. Finally, we will indicate whether our method can capture language-specific knowledge on the basis of retaining general knowledge.  
%In Appendix~\ref{variet}, we verify our method better than different varieties
%In the analysis section, we will first report the importance situation of different neurons and languages, then analyze the distribution of different neurons and modules in the model, then show the results of different specific capacity, and next verify our method better than different varieties. Finally, we will indicate whether our method can capture language-specific knowledge on the basis of retaining general knowledge.

%\subsection{Distribution of the Language Importance}

\subsection{Neuron Importance for Different languages}
In our method, we allocate neurons based on their importance for different languages. The rationality behind this mechanism is that different neurons should have distinct importance values so that these neurons can find their relevant language pairs. Therefore, we show the importance of neurons computed by Taylor Expansion in different modules for the one-to-many (O2M) and many-to-many (M2M) translation tasks. For clarity and convenience, we only show the importance values of three language pairs in the sixth layer of encoder and decoder. %These language pairs are It$\rightarrow$En, Ro$\rightarrow$It, and En$\rightarrow$Ro, which are 0.67, 1, and 1.7 higher than the multilingual baseline in bleu, respectively. 

The results of O2M are shown in Figure~\ref{importance1} and Figure~\ref{importance2}, and the language pairs are En$\rightarrow$Es, En$\rightarrow$Pt, and En$\rightarrow$Fi. The first two target languages are Spanish and Portuguese, both of which belong to the Western Romance, the Romance branch of the Indo-European family, while the last one is Finnish, a member of the Finnish-Ugric branch of the Ural family. As we can see, the importance of Spanish and Portuguese are always similar in most neurons, but there is no obvious correlation between Finnish and the other two languages. It indicates that similar languages are also similar in the distribution of the neuron importance, which implies that the common features in similar languages can be captured by the same neurons.
%and how we can allocate neurons based on linguistic knowledge and relationships.

The results of M2M are shown in Figure~\ref{importance3} and Figure~\ref{importance4}, and the language pairs are It$\rightarrow$En, Ro$\rightarrow$It, and En$\rightarrow$Ro, whose BLEU scores are 0.67, 1, and 1.7 higher than the multilingual baseline, respectively. In most neurons, the highest importance value is twice as high as the lowest and this high variance of importance provides the theoretical basis for later neuron allocation. Moreover, we can see a lot of importance peaks of the two language pairs: Ro$\rightarrow$It and En$\rightarrow$Ro, which means that these neurons are especially important for generating the translation for these language pairs. However, the fluctuation of It$\rightarrow$En is flat with almost no peaks, which means only a few neurons are specific to this language pair. This may be the reason why some language pairs have higher improvements, while some have lower improvements.%translation performance 

\subsection{Distribution of the Language-specific Neurons}

Except for the general neurons shared by all the language pairs, our method allocates other neurons to different language pairs based on their importance. These language-specific neurons are important for preserving the language-specific knowledge. To better understand the effectiveness of our method, we will show how these specific neurons are distributed in the model.
%We next study how allocation schedules this specific capacity across all Transformer sub-layers, in order to determine the ideal arrangement of language-specific neurons.

To evaluate the proportion of language-specific neurons for different language pairs at each layer, we introduce a new metric, LScore, formulated as:
\begin{equation}
    \mathrm { LScore }(l, m)=\frac{\tilde{I}_{l}^{m}}{\tilde{I}_l}, m \in \{1,\ldots, M\}
\label{lscore}
\end{equation}
where $\tilde{I}_{l}^{m}$ denotes the number of neurons allocated to language pair $m$ in the $l$-th layer, and $\tilde{I}_l$ denotes the total number of the language-specific neurons in the $l$-th layer. The larger the LScore, the more neurons allocated to the language pair $m$.
We also introduce a metric to evaluate the average proportion of language-specific neurons of each language in different modules, which formulated as:
%proportion of neurons in different module is specific, MScore , formulated as follows:
\begin{equation}
    \mathrm{MScore}(l, f)=\frac{1}{M} \sum_{m=0}^{M} \frac{\tilde{I}_{l, f}^{m}}{\tilde{I}_{l,f}}, m \in \{1,\ldots, M\}
\label{lscore}
\end{equation}
where $\tilde{I}_{l, f}^{m}$ denotes the number of specific neurons for language pair $m$ of in the $f$ module of the $l$-th layer and $M$ denotes the total number of the language pair. The larger the MScore is, the more specific neurons are allocated to different language pairs in this module.
%A larger LScore or MScore indicates that this sub-layer or module occupy more neurons, i.e., each neuron allocate to more language pairs and it is more sharing. On the contrary, a smaller LScore or MScore indicates that it is more specific. 

As shown in Figure~\ref{sumenc} and Figure~\ref{sumdec}, the language pairs have low LScores at the top and bottom layers and high LScores at the middle layers of both the encoder and decoder. The highest LScore appears at the third or fourth layers, which indicates that the neuron importance of different language pairs is similar and the neurons of the middle layers are shared by more languages. As a contrast, the bottom and top layers will be more specialized for different language pairs. Next, from Figure~\ref{2enc} and Figure~\ref{3dec}, we can see the MScores of the attention modules are almost near 1.0, which means neurons in self attention and cross attention are almost shared across all language pairs. However, the MScores of Feed Forward Network (FFN) gradually decrease as layer depth increases and it shows that the higher layers in FFN are more essential for capturing the language-specific knowledge.

\subsection{Effects of the Hyper-parameter $k$}
\label{sec:kmax}
%\subsection{Proportion of the Specific Capacity}
% the range of k

%%%%%%%%%%%%%%the range of k%%%%%%%%%%%%%
\begin{figure}[t!]
    \centering
    \includegraphics[width=0.95\columnwidth]{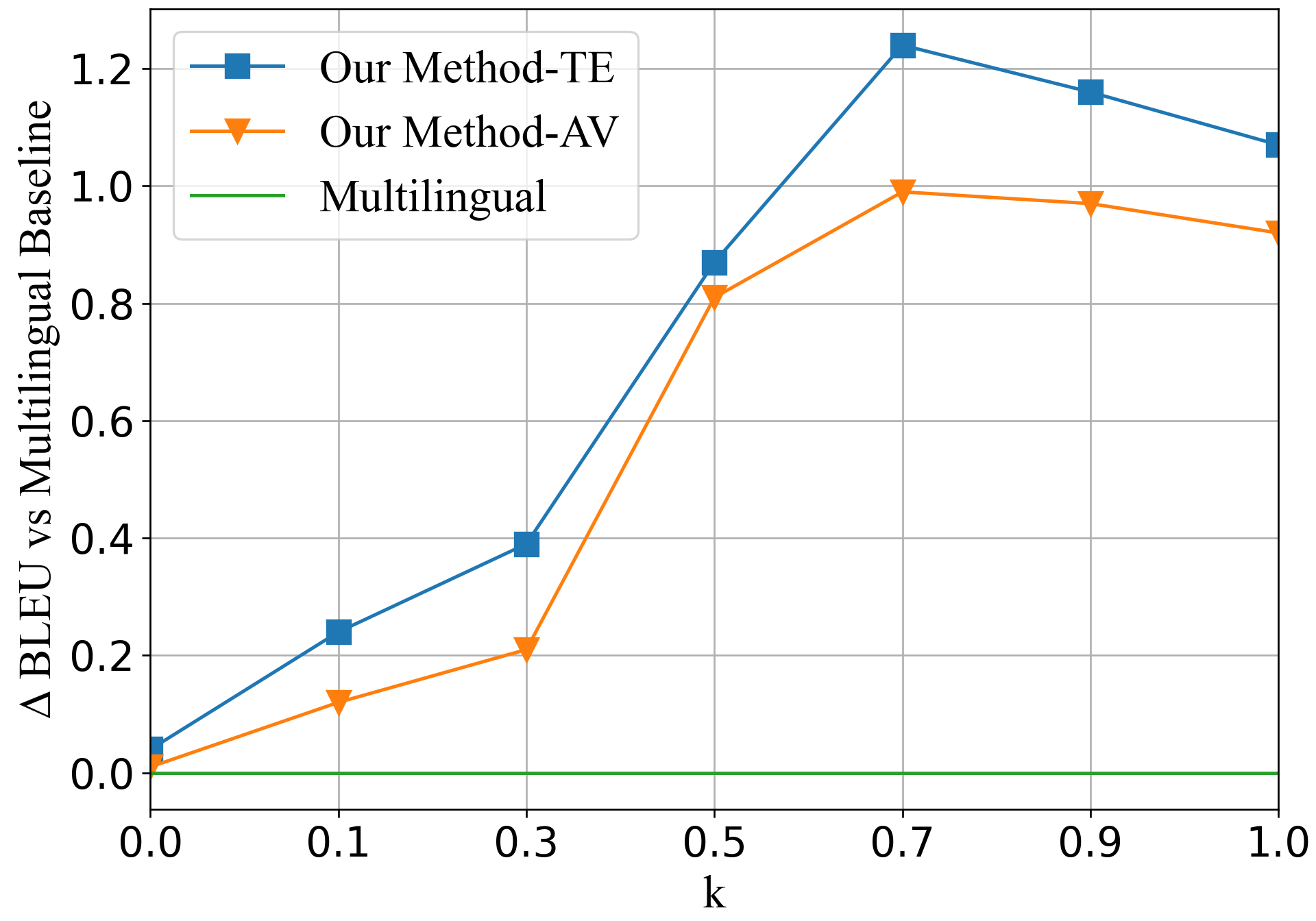}
    \caption{The average $\Delta$ BLEU over the Multilingual baseline  with different hyper-parameters $k$ on the many-to-many translation task.}
    \label{fig:krange}
\end{figure}

When the importance of neurons for different languages is determined, the number of language pairs associated with each neuron can be adjusted according to $k$.
%As we adopt a hierarchical sharing mode in language-specific neurons, we should conduct experiments to evaluate translation performance under different sharing degree.
When $k=1.0$, the threshold is $\max(\Theta^m(i))$ as computed by Equation~\ref{eq:kmax}, so the neurons will only be allocated to the language pair with the highest importance, and when $k=0$, the threshold is 0 so the neurons will be shared across all language pairs just like the Multilingual baseline. To better show the overall impact of the hyper-parameter $k$, we vary it from $0$ to $1$ and the results are shown in Figure~\ref{fig:krange}. As we can see, the translation performance of the two proposed approaches increases with the increment of $k$ and reach the best performance when $k$ equals $0.7$. As $k$ continues to increase, the performance deteriorates, which indicates that the over-specific neurons are bad at capturing the common features shared by similar languages and will lead to performance degradation.

%more specific allocation fails to deliver increasingly better translation performance.

%%%%%%%%%%%general and specific performance%%%%%%%%%%%%
\begin{figure}[t!]
    \centering
    \includegraphics[width=\columnwidth]{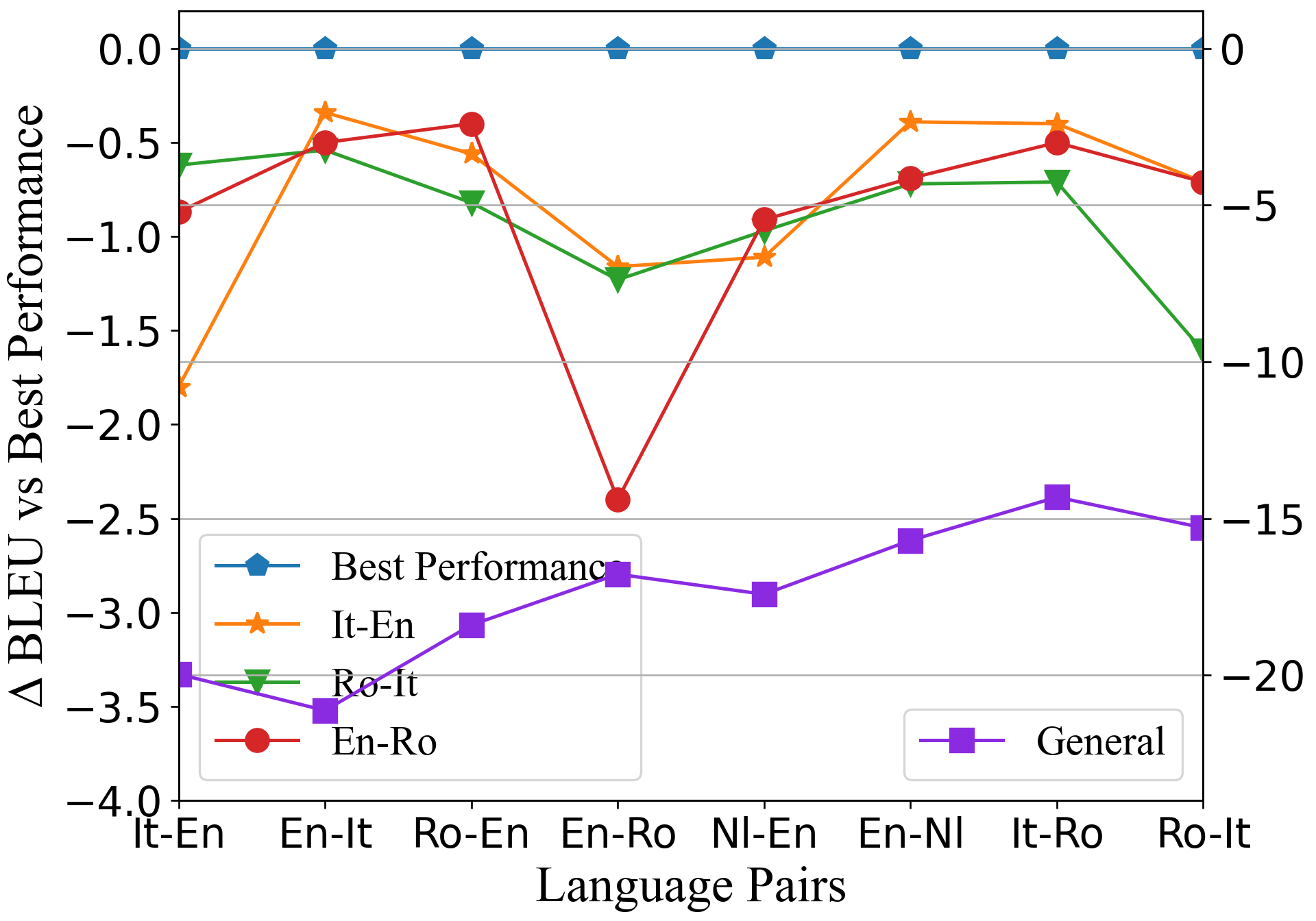}
    \caption{$\Delta$ BLEU over best performance when erasing the general or language-specific neurons randomly on the many-to-many translation task.}
    \label{fig:general}
\end{figure}

\subsection{The Specific and General knowledge}

The main idea of our method is to let the general knowledge and the language-specific knowledge be captured by different neurons of our method. 
%we assess whether this achieved its goal, i.e. did the two types of neurons we allocated perform their respective functions? 
To verify whether this goal has been achieved, we conduct the following experiments.
For the general knowledge, we randomly erase 20\% general neurons of the best checkpoint of our method, which means we mask the output value of these neurons to $0$, then generate translation using it. For language-specific knowledge, we randomly erase 50\% specific neurons and then generate translation.

As shown in Figure~\ref{fig:general}, when the general neurons are erased, the BLEU points of all the language pairs drop a lot (about 15 to 20 BLEU), which indicates general neurons do capture the general knowledge across languages. For specific neurons, we show three language pairs for the sake of convenience. We can see that when the neurons associated with the current language pair are erased, the performance of this language pair decreases greatly. However, the performance of other language pairs only declines slightly, because the specific knowledge captured by these specific neurons are not so important for other languages.
%some erased neurons may also be specific neurons to other language pairs according to our neuron allocation method.

\section{Related Work}

Our work closely relates to language-specific modeling for MNMT and model pruning which we will recap both here. Early MNMT studies focus on improving the sharing capability of individual bilingual models to handle multiple languages, which includes sharing encoders~\citep{DBLP:conf/acl/DongWHYW15}, sharing decoders~\citep{DBLP:conf/emnlp/ZophYMK16}, and sharing sublayers~\citep{DBLP:conf/naacl/FiratCB16}. Later, ~\citet{DBLP:journals/corr/HaNW16} and ~\citet{DBLP:journals/tacl/JohnsonSLKWCTVW17} propose an universal MNMT model with a target language token to indicate the translation direction. While this paradigm fully explores the general knowledge between languages and hard to obtain the specific knowledge of each language~\citep{DBLP:conf/emnlp/TanCHXQL19,DBLP:conf/naacl/AharoniJF19}, the subsequent researches resort to Language-specific modeling, trying to find a better trade-off between sharing and specific. Such approaches involve inserting conditional language-specific routing layer~\citep{zhang2021share}, specific attention networks~\citep{DBLP:conf/coling/BlackwoodBW18,DBLP:conf/wmt/SachanN18}, adding task adapters~\citep{DBLP:conf/emnlp/BapnaF19}, and training model with different language clusters~\citep{DBLP:conf/emnlp/TanCHXQL19}, and so on. However, these methods increase the capacity of the model which makes the model bloated.

Moreover, our method is also related to model pruning, which usually aims to reduce the model size or improve the inference efficiency. Model pruning has been widely investigated for both computer vision (CV)~\citep{DBLP:conf/iccv/LuoWL17} and natural language processing (NLP) tasks. For example, \citet{DBLP:conf/conll/SeeLM16} examines three magnitude-based pruning schemes, \citet{DBLP:conf/iclr/ZhuG18} demonstrates that large-sparse models outperform comparably-sized small-dense models, and \citet{DBLP:conf/emnlp/WangWLT20} improves the utilization efficiency of parameters by introducing a rejuvenation approach. Besides, \citet{DBLP:conf/iclr/LanCGGSS20} presents two parameter reduction techniques to lower memory consumption and increase the training speed of BERT. \citet{DBLP:conf/naacl/GuFX21} prune then expand the model neurons or parameters based on importance on domain adaptation of neural machine translation.

\section{Conclusion}

The current standard models of multilingual neural machine translation fail to capture the characteristics of specific languages, while the latest researches focus on the pursuit of specific knowledge while increasing the capacity of the model and requiring fine manual design. To solve the problem, we propose an importance-based neuron allocation method. We divide neurons to general neurons and language-specific neurons to retain general knowledge and capture language-specific knowledge without model capacity incremental and specialized design. The experiments prove that our method can get superior translation results with better general and language-specific knowledge.

\section*{Acknowledgments}

We thank all the anonymous reviewers for their insightful and valuable comments. This work was supported by National Key R\&D Program of China (NO. 2017YFE0192900).

\bibliographystyle{acl_natbib}
\bibliography{anthology,acl2021}

\clearpage

\appendix

\begin{table*}[t!]
\centering
\resizebox{2.0\columnwidth}!{
\begin{tabular}{llllllllll}
\bottomrule
 & It$\rightarrow$En & En$\rightarrow$It & Ro$\rightarrow$En & En$\rightarrow$Ro & Nl$\rightarrow$En & En$\rightarrow$Nl & It$\rightarrow$Ro & Ro$\rightarrow$It & AVE \\ 
\hline
\hline
\textbf{$\rho = 80\%$} & 38.3 & 34.05 & 32.11 & 26.01 & 32.24 & 30.12 & 21.96 & 24.39 & 29.94 \\
\textbf{$\rho = 90\%$} & \textbf{38.31} & \textbf{34.15} & \textbf{32.24} & \textbf{26.34} & \textbf{32.73} & \textbf{30.16} & \textbf{22.21} & \textbf{24.76} & \textbf{30.11} \\
\textbf{$\rho = 95\%$} & 38.28 & 33.82 & 32.05 & 25.74 & 31.97 & 29.51 & 21.56 & 24.19 & 29.64 \\
\bottomrule
\end{tabular}
}
\caption{BLEU scores on many-to-many translation tasks when $k=0.7$}
\label{tab:rho}
\end{table*}

\section{Performance on Different Varieties}
\label{variet}

\begin{figure}
    \centering
    \includegraphics[width=\columnwidth]{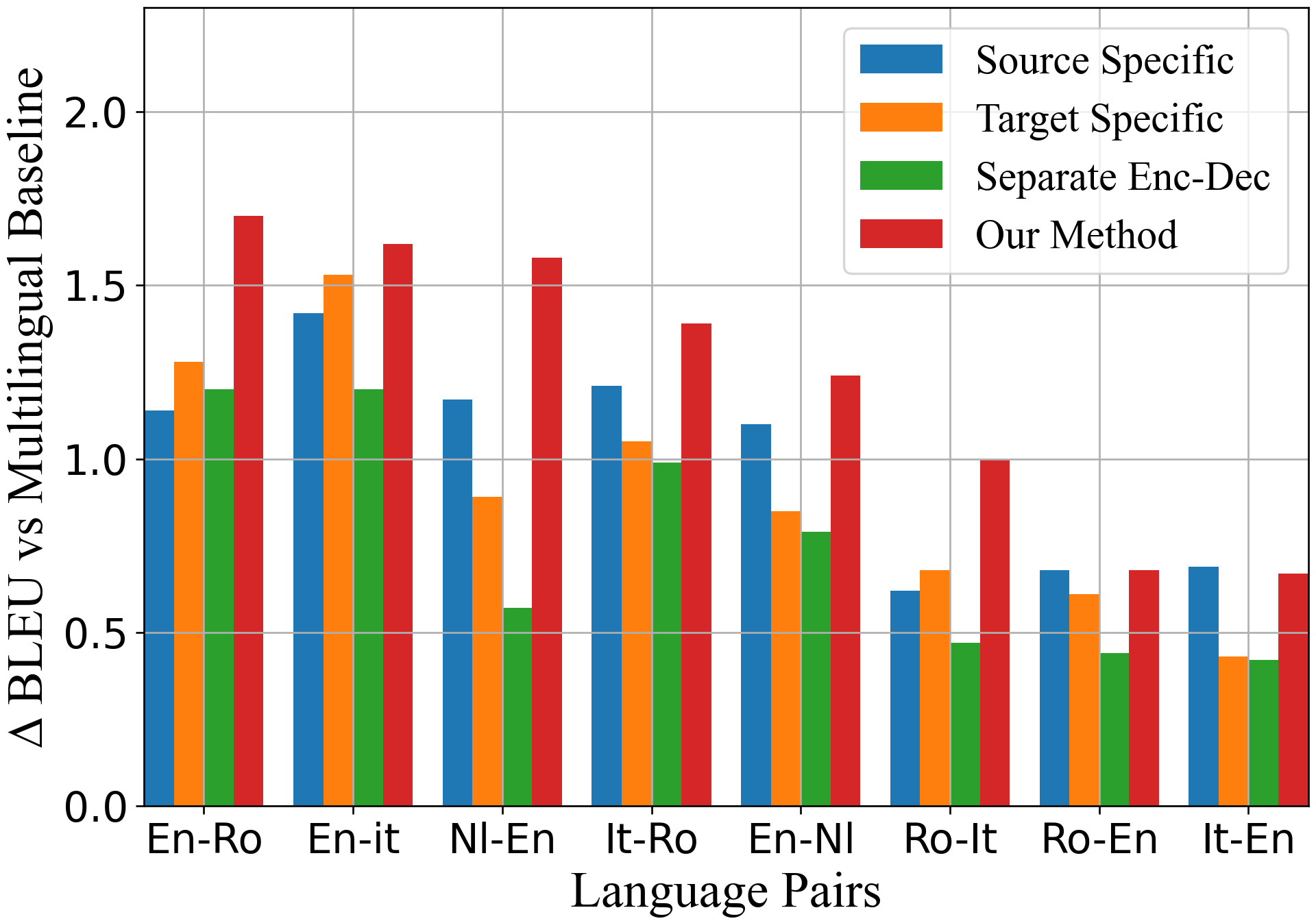}
    \caption{$\Delta$ BLEU over Multilingual baseline on many-to-many translation.}
    \label{fig:encdec}
\end{figure}

In the proposed method we allocate neurons based on importance of language pair. There are three varieties of our method: (a) Source-Specific, share all neurons according to the source language only; (b) Target-Specific, share all neurons according to the target language only; (c) Separate Enc-Dec, Encoder neurons are shared according to the source language and decoder neurons are shared according to the target language. Note that (c) is different from our method since (c) is separate neurons to two parts (encoder and decoder) and then connect specific neurons of the two parts to form a whole, while our method is directly based on language pairs.

As shown in Figure~\ref{fig:encdec}, we compare our Taylor Expansion method with the other three varieties. Our approach outperforms other varieties on almost all language pairs, and the performance of the language-pair based approach is undoubtedly the best. The second is based on the target language and the source language. Worst of all are the separated encoder-decoder, which may be due to the mismatch between the neurons of the encoder and decoder when they are reconnected.

\section{Effects of the Hyper-parameter $\rho$}
\label{app:rho}
We conducted several experiments on $\rho$ to determine the optimal hyper-parameter, so as to determine the proportion of universal neurons. As shown in Table~\ref{tab:rho}, when $\rho = 90\%$ the model gets the best translation result and reach best trade-off between general and language-specific neurons.

\end{document}